\documentclass[conference]{IEEEtran}
\ifCLASSINFOpdf
\else
\fi

\usepackage{graphicx}
\usepackage{color,soul}
\usepackage{amsmath, amsthm, amssymb}
\usepackage{amsmath}
\usepackage{framed}
\usepackage{float}
\usepackage{subcaption}
\usepackage{url}
\usepackage{fancyhdr}
\usepackage{algorithm}
\usepackage{pifont}
\usepackage[noend]{algpseudocode}
\usepackage{listings}
\usepackage{color}
\usepackage{tikzpagenodes}

\usepackage[textsize=tiny]{todonotes}

\begin{document}

%\title{Representing Big Data Knowledge using both Vector Spaces and Knowledge Graphs}
%\title{Thinking Fast, Thinking Slow! Combining Knowledge Graphs and Vector Spaces}

\title{Thinking, Fast and Slow:\\Combining Vector Spaces and Knowledge Graphs}


\author{\IEEEauthorblockN{Sudip Mittal, Anupam Joshi, Tim Finin}
\IEEEauthorblockA{University of Maryland, Baltimore County, Baltimore, MD 21250, USA\\ Email: $\lbrace$smittal1,joshi,finin$\rbrace$@umbc.edu}
}


%Telephone: (800) 555--1212, Fax: (888) 555--1212}


\maketitle

\begin{abstract}

Knowledge graphs and vector space models are robust knowledge representation techniques with individual strengths and weaknesses. Vector space models excel at determining similarity between concepts, but are severely constrained when evaluating complex dependency relations and other logic-based operations that are a strength of knowledge graphs. We describe the \textit{VKG structure} that helps unify knowledge graphs and vector representation of entities, and enables powerful inference methods and search capabilities that combine their complementary strengths. We analogize this to thinking `fast' in vector space along with thinking `slow' and `deeply' by reasoning over the knowledge graph. We have created a query processing engine that takes complex queries and decomposes them into subqueries optimized to run on the respective knowledge graph or vector view of a VKG. We show that the VKG structure can process specific queries that are not efficiently handled by vector spaces or knowledge graphs alone. We also demonstrate and evaluate the VKG structure and the query processing engine by developing a system called \textit{Cyber-All-Intel} for knowledge extraction, representation and querying in an end-to-end pipeline grounded in the cybersecurity informatics domain. %The system creates a cybersecurity corpus by collecting threat and vulnerability intelligence from various textual sources like, national vulnerability databases, dark web vulnerability markets, social networks, blogs, etc. These are represented as instances of our VKG structure. We then use the system to answer complex cybersecurity informatics queries. 

\end{abstract}


% \fi
\IEEEpeerreviewmaketitle

\section{Introduction}\label{intro}
Knowledge representation is one of the central challenges in the field of Artificial Intelligence. It involves designing computer representations that capture information about the world that can be used to solve complex problems \cite{davis1993knowledge}. We introduce the \textit{VKG structure} as a hybrid structure that combines knowledge graph and embeddings in a vector space. The structure creates a new representation for relations and entities of interest. In the VKG structure, the knowledge graph includes explicit information about various entities and their relations to each other grounded in an ontology schema. The vector embeddings, on the other hand, include implicit information found in context where these entities occur in a corpus \cite{mikolov2013efficient,pennington2014glove}.

While representing knowledge through different methods there is an inherent information loss. For example, by representing knowledge as just vector embeddings we lose the declarative character of the information. While knowledge graphs are adept at asserting declarative information, they miss important contextual information around the entity or are restricted by the expressibility of the baseline ontology schema used to represent the knowledge \cite{davis1993knowledge}. 

In discussing the pros-and-cons of knowledge graphs and vector space models, it's important to highlight the fact that both of the knowledge representation techniques provide applications built on these technology's advantages. Embeddings provide an easy way to search their neighborhood for similar concepts and can be used to create powerful deep learning systems for specific complex tasks. Knowledge graphs provide access to versatile reasoning techniques. Knowledge graphs also excel at creating rule-based systems where domain expertise can be leveraged.

To overcome limitations of both and take advantage of their complimentary strengths, we have developed the VKG structure that is part knowledge graph and part vector embeddings. Together, they can be used to develop powerful inference methods and a better semantic search. 

We present our method to create the VKG structure from a given textual corpus and discuss special types of queries that can take advantage of this structure, specifically the queries that are decomposed automatically into sub-queries that run efficiently on the vector space part and the knowledge graph part. We also discuss some use-cases and applications that can take advantage of the VKG structure.

This approach is reminiscent of Daniel Kahneman's model of the human cognitive systems that was the topic of his popular book {\em Thinking, Fast and Slow} \cite{kahneman2011}.  He posited two cognitive mechanisms, a {\em System 1} that is fast and instinctive and a {\em System 2} that is slow, deliberative and more logical.

To demonstrate and evaluate the VKG structure and the query processing engine we have developed a system called \textit{Cyber-All-Intel}. Our system strives to bring in ``intelligence" about the state of the cyber world, such as an analyst might obtain by looking, for instance, at security blogs, NVD/CVE updates, and parts of the dark web. It represents this cybersecurity informatics knowledge in the VKG structure. The system can answer complex cybersecurity informatics queries and process tasks like, \textit{`Issue an alert if a vulnerability similar to denial of service is found in MySQL'}, which involves various operations on the VKG structure, including searching for similar vulnerabilities, querying to find known vulnerabilities in a product, and reasoning to come up with an alert.

\vspace{2mm}

The main contributions of this paper are: 
\begin{itemize}
\item The \textit{VKG structure} that is part vector embeddings and part knowledge graph. 
\item A query processing engine that decomposes an input query to $search$, $list$, and $infer$ on the VKG structure. 
\item A cybersecurity informatics system built on top of the VKG structure called \textit{Cyber-All-Intel}. 
\end{itemize}
In the rest of the paper we discuss the related work, describe the details of the VKG structure, present an evaluation, and make concluding remarks.

\section{Related Work}\label{relwork}
In this section we present some related work on knowledge graphs, vector space models, and knowledge representation for cybersecurity.
%Our proposed work will add support for streaming data to a cybersecurity knowledge base. Modifying static knowledge bases to handle dynamic data will help us in reasoning over dynamic and updating data. Also, in our approach we aim to integrate word embeddings and entity vectors with our knowledge graph which makes responses to certain known analyst tasks faster as described next.
\subsection{Vector Space Models \& Knowledge Graphs}

%Vector-based representation of textual data has become increasingly popular in natural language processing (NLP) and cognitive science. 
%Compositional Vector Space Models have been developed to represent phrases and sentences in natural language as vectors where various frameworks are used for representing the meaning of phrases and sentences in vector space. \cite{mitchell2008vector,baroni2010nouns}. 

Word embeddings are used to represent words in a continuous vector space. Two popular methods to generate these embeddings based on `Relational Concurrence Assumption' are word2vec \cite{mikolov2013distributed,mikolov2013efficient} and GloVe \cite{pennington2014glove}. The main idea behind generating embeddings for words is to say that vectors close together are semantically related. Word embeddings have been used in various applications like machine  translation \cite{sutskever2014sequence}, improving local and global context \cite{huang2012improving}, etc. 

Modern knowledge graphs assert facts in the form of ($Subject$, $Predicate$, $Object$) triples, where the $Subject$ and $Object$ are modeled as graph nodes and the edge between them ($Predicate$) model the relation between the two. DBpedia \cite{auer2007dbpedia}, YAGO \cite{suchanek2007yago}, YAGO2 \cite{hoffart2013yago2}, and the Google Knowledge Graph \cite{officialgoogle2012} are examples of popular knowledge graphs. The underlying representation languages of some knowledge graphs (e.g., DBpedia, Freebase \cite{bollacker2008freebase}) support including rich schema-level knowledge and axioms along with the instance-level graph.

An important task for vector space models and knowledge graphs is finding entities that are similar to a given entity. In vector spaces, embeddings close together are semantically related and various neighborhood search algorithms \cite{gionis1999similarity,Kuzi:2016:QEU:2983323.2983876} are used for this task. On the other hand semantic similarity on knowledge graphs using ontology matching, ontology alignment, schema matching, instance matching, similarity search, etc. remains a challenge \cite{shvaiko2013ontology,DeVirgilio:2013:SMA:2457317.2457352,zheng2016semantic}. Sleeman et al. \cite{sleeman2015topic} used vectors based on topic models to compare the similarity of nodes in RDF graphs. In this paper we propose the VKG structure, in which we link the knowledge graph nodes to their embeddings in a vector space (see Section \ref{model}).

%\textbf{Todo add 2 papers on kg semantic similarity}% vldb16_sed }%http://dl.acm.org/citation.cfm?id=2457352}

%Kasneci et al. \cite{kasneci2008naga} and Ding et al. \cite{ding2004swoogle} created a semantic search engine. 
Yang et al. \cite{yang2016fast} argued that a fast top-k search in knowledge graphs is challenging as both graph traversal and similarity search are expensive. The problem will get compounded as knowledge graphs increase in size. Their work proposes STAR, a top-k knowledge graph search framework to find top matches to a given input. Janovic et al. \cite{damljanovic2011random} have suggested using Random Indexing (RI) to generate a semantic index to an RDF\footnote{\url{https://www.w3.org/RDF/}} graph. These factors combined have led to an increased interest in semantic search, so as to access RDF data using Information Retrieval methods. We argue that vector embeddings can be used to search, as well as index entities in a knowledge graph. We have built a query engine on top of the VKG structure that removes the need to search on the knowledge graph and uses entity vector embeddings instead (see Section \ref{model} and \ref{eval}). However, queries that involve listing declarative knowledge and reasoning are done on the knowledge graph part of the VKG structure. 

%vectors + KG

Vectorized knowledge graphs have also been created, systems like HOLE (holographic embeddings) \cite{nickel2016holographic} and TransE \cite{wang2014knowledge} learn compositional vector space representations of entire knowledge graphs by translating them to different hyperplanes. Our work is different from these models as we keep the knowledge graph part of the VKG structure as a traditional knowledge graph so as to fully utilize mature reasoning capabilities and incorporate the dynamic nature of the underlining corpus for our cybersecurity use-case. Vectorizing the entire knowledge graph part for a system like Cyber-All-Intel will have significant computational overhead because of the ever-changing nature of vulnerability relations and velocity of new input threat intelligence.

In another thread different from ours, vector models have been used for knowledge graph completion. Various authors have come up with models and intelligent systems to predict if certain nodes in the knowledge graphs should have a relation between them. The research task here is to complete a knowledge graph by finding out missing facts and using them to answer path queries \cite{lin2015learning,neelakantan2015compositional,socher2013reasoning,guu2015traversing}.

\subsection{Knowledge Representation for Cybersecurity}

Knowledge graphs have been used in cybersecurity to combine data and information from multiple sources. Undercoffer et al. \cite{Undercoffer2003b} developed an ontology by combining various taxonomies for intrusion detection. Kandefer et al. \cite{kandefer2007symbolic} created a data repository of system vulnerabilities and with the help of a systems analyst, trained a system to identify and prevent system intrusion. Takahashi et al. \cite{takahashi2010ontological,takahashi2010building} built an ontology for cybersecurity information based on actual cybersecurity operations focused on cloud computing-based services.  Rutkowski et al. \cite{rutkowski2010cybex} created a cybersecurity information exchange framework, known as CYBEX. The framework describes how cybersecurity information is exchanged between cybersecurity entities on a global scale and how implementation of this framework will eventually minimize the disparate availability of cybersecurity information. % Another insightful work by Xie et al. \cite{xie2010using} discusses uncertainty modeling for cyber security centered around near real-time security analysis such as intrusion response. In this paper the authors use Bayesian networks to model uncertainty in enhanced security analysis.

Syed et al. \cite{syed2015uco} created the Unified Cybersecurity Ontology (UCO) that supports information integration and cyber situational awareness in cybersecurity systems. The ontology incorporates and integrates heterogeneous data and knowledge schema from different cybersecurity systems and most commonly-used cybersecurity standards for information sharing and exchange such as STIX and CYBEX \cite{rutkowski2010cybex}. The UCO ontology has also been mapped to a number of existing cybersecurity ontologies as well as concepts in the Linked Open Data cloud.

Various preliminary systems \cite{Joshi-ICSC-2013,mittal2016cybertwitter} demonstrate the feasibility of automatically generating RDF linked data from vulnerability descriptions collected from the National Vulnerability Database\footnote{\url{https://nvd.nist.gov/}}, Twitter,  and other sources. The system by Joshi et al. \cite{Joshi-ICSC-2013} extracts information on cybersecurity-related entities, concepts and relations which is then represented using custom ontologies for the cybersecurity domain and mapped to objects in the DBpedia knowledge base \cite{auer2007dbpedia} using DBpedia Spotlight \cite{mendes2011dbpedia}. Mittal et al. \cite{mittal2016cybertwitter} developed CyberTwitter, a framework to automatically issue cybersecurity vulnerability alerts to users. CyberTwitter converts vulnerability intelligence from tweets to RDF. It uses the UCO ontology (Unified Cybersecurity Ontology) \cite{syed2015uco} to provide their system with cybersecurity domain information.

%For an intelligent system like CyberTwitter, it is vital to understand the difference between various real world concepts and also to posses a comprehensive knowledge about the cybersecurity domain. In this paper we use various publicly available cybersecurity ontologies and knowledge bases to support information integration and cyber-situational awareness: 

%\begin{enumerate}
%\item UCO: Unified Cybersecurity Ontology~\cite{syed2015uco}: The ontology integrates heterogeneous data and knowledge schemas from different cybersecurity systems and standards.
%\item DBpedia\cite{auer2007dbpedia}: DBpedia is a project to extract structured content from the information created as part of the Wikipedia project\footnote{\url{https://wikimediafoundation.org/wiki/Our_projects}}. 
%\item YAGO (Yet Another Great Ontology)~\cite{suchanek2007yago}: It is a knowledge base automatically extracted from Wikipedia and other sources.
%\end{enumerate}

\section{VKG Structure}\label{model}
\begin{figure}[h]
\centering
\includegraphics[scale=0.18]{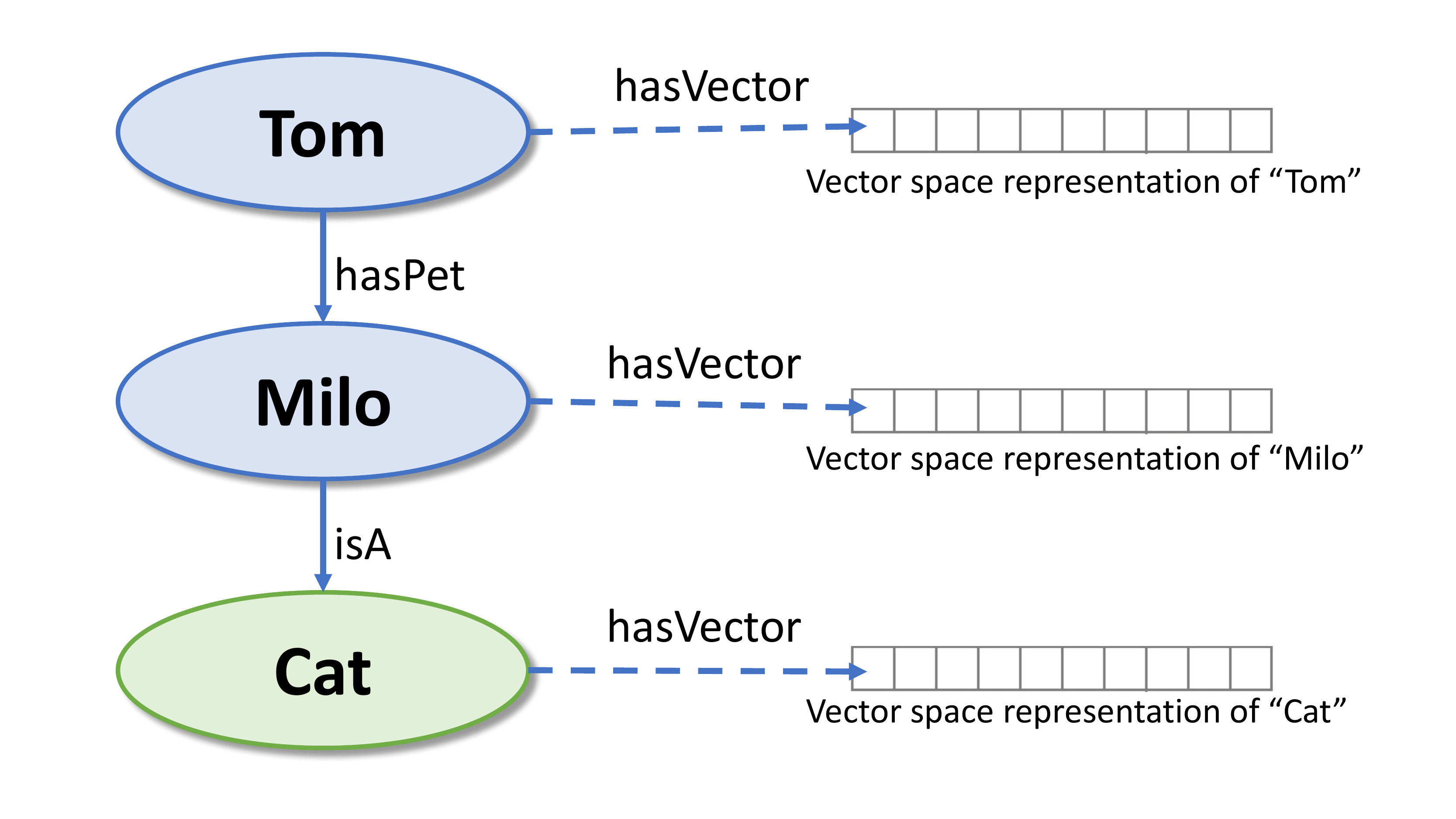}
\caption{An example of a VKG structure representing ``Tom has a pet Milo and Milo is a cat''.}
\label{fig:ex}
\end{figure}

In this section we describe our VKG structure, which leverages both vector spaces and knowledge graphs to create a new representation for relations and entities of interest present in text. In the VKG structure, an entity is represented as a node in a knowledge graph and is linked to its representation in a vector space. Figure \ref{fig:ex} gives an example of the VKG structure where entity nodes are linked to each other using explicit relations as in a knowledge graph and are also linked to their word embeddings in a vector space. The VKG structure enables an application to reason over the knowledge graph portion of the structure and also run computations on the vector space part. 

The VKG structure enables us to specifically assert information present in the vector representation of concepts and entities using semantic relations, for example, in Figure \ref{fig:ex}; using the VKG structure we can explicitly assert that the vector representation of `Milo' and `Cat' are related like, $<Milo, isA, Cat>$. `Tom' and `Milo' are related with the relation, $<Tom, hasPet, Milo>$. The vector representation of `Milo' can be used to find other pets that are similar to it, but the explicit information that `Milo' is a `Cat' and it's `Tom' that has a pet named `Milo' can be accurately derived explicitly from the knowledge graph part. 

The VKG structure helps us unify knowledge graphs and vector representation of entities, and allows us to develop powerful inference methods that combine their complementary strengths.

The vector representation we use is based on the `Relational Concurrence Assumption' highlighted in \cite{mikolov2013distributed,mikolov2013efficient,collobert2011natural}. Word embeddings are able to capture different degrees of similarity between words. Mikolov et al. \cite{mikolov2013distributed,mikolov2013efficient} argue that embeddings can reproduce semantic and syntactic patterns using vector arithmetic. Patterns such as ``Man is to Woman as King is to Queen'' can be generated through algebraic operations on the vector representations of these words such that the vector representation of ``King'' - ``Man'' + ``Woman'' produces a result which is closest to the vector representation of ``Queen'' in the model. Such relationships can be generated for a range of semantic relations as well as syntactic relations. However, in spite of the fact that vector space models excel at determining similarity between two vectors they are severely constrained while creating complex dependency relations and other logic based operations that are a forte of various semantic web based applications \cite{davis1993knowledge,berners2001semantic}. 

Knowledge graphs, on the other hand, are able to use powerful reasoning techniques to infer and materialize facts as explicit knowledge. Those based on description logic representation frameworks like OWL, for example, can exploit axioms implicit in the graphs to compute logical relations like consistency, concept satisfiability, disjointness, and subsumption. As a result, they are generally much slower while handling operations like, ontology alignment, instance matching, and semantic search \cite{shvaiko2013ontology,JeanMary2009OntologyMW}.

The intuition behind our approach is that an entity's context from its immediate neighborhood, present as word embeddings, adds more information along with various relations present explicitly in a knowledge graph. Entity representation in vector space gives us information present in the immediate context of the place where they occur in the text and knowledge graphs give us explicit information that may or may not be present in the specific piece of text. Embeddings can help find similar nodes or words faster using neighborhood search algorithms and search space reductions. They also support partial matching techniques. Knowledge graphs provide many reasoning tools including query languages like SPARQL\footnote{\url{https://www.w3.org/TR/rdf-sparql-query/}}, rule languages like SWRL\footnote{\url{https://www.w3.org/Submission/SWRL/}}, and description logic reasoners.

Potential applications that will work on our VKG structure, need to be designed to take advantages provided by integrating vector space models with a knowledge graph. In a general efficient use-case for our VKG structure, `fast' top-k search should be done on the vector space part aided by the knowledge graph, and the `slow' reasoning based computations should be performed on just the knowledge graph part. An input query can be decomposed into sub-queries which run on respective parts of the VKG structure (see \ref{query}). We analogize this to thinking `fast' in vector space along with thinking `deeply' and `slowly' by using the knowledge graph. 

%\begin{figure}[h]
%\centering
%\includegraphics[scale=0.19]{images/domaininformation.pdf}
%\caption{Adding medical domain knowledge using the VKG structure.}
%\label{fig:domain}
%\end{figure}

\begin{figure*}[ht]
\centering
\includegraphics[scale=0.28]{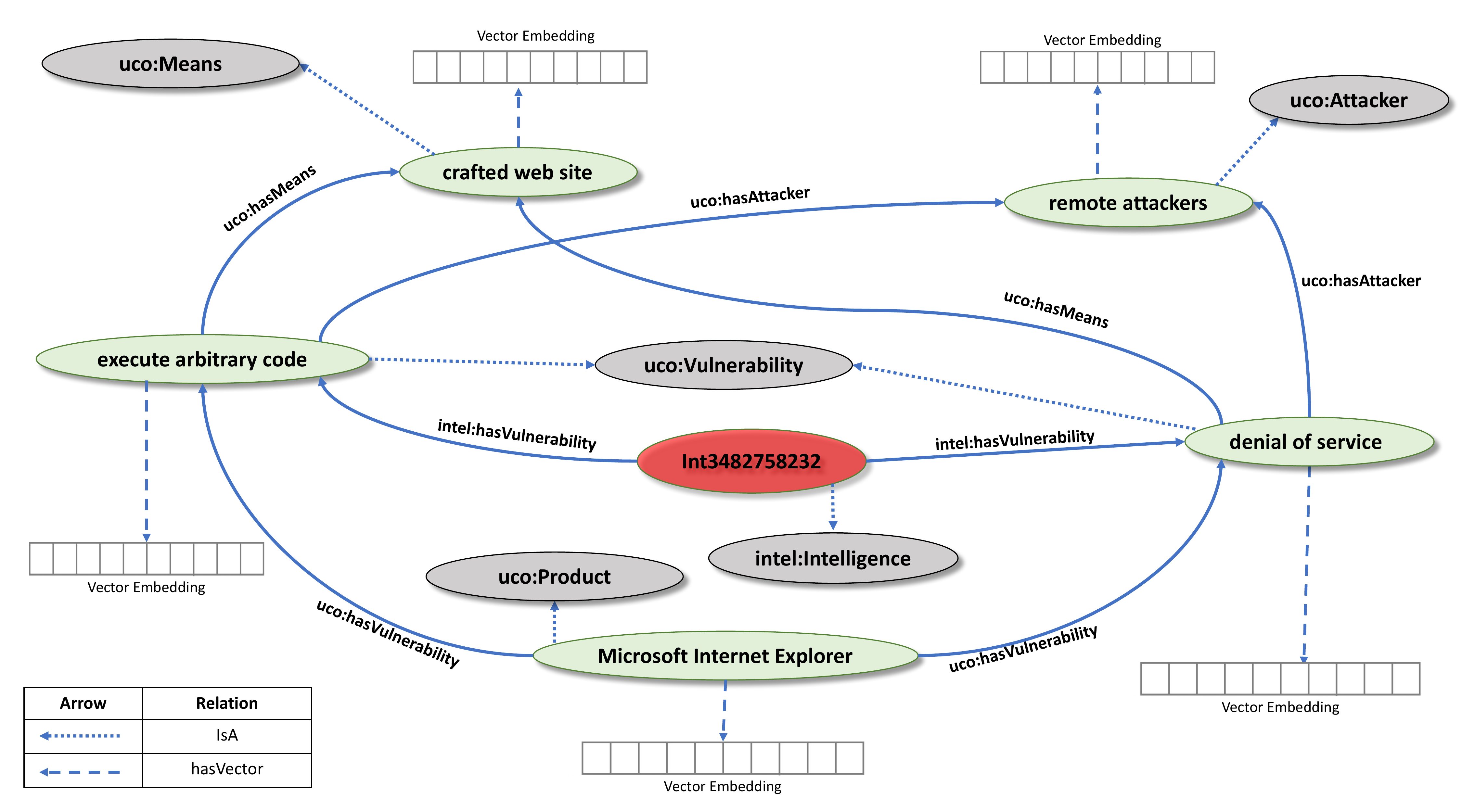}
\caption{In the VKG structure for \textit{``Microsoft Internet Explorer allows remote attackers to execute arbitrary code or cause a denial of service (memory corruption) via a crafted web site, aka ``Internet Explorer Memory Corruption Vulnerability.''} the knowledge graph part asserted using UCO includes the information that a product `Microsoft Internet Explorer' has vulnerabilities `execute arbitrary code' and `denial of service' that can be exploited by `remote attackers' using the means `crafted web site'. The knowledge graph entities are linked to their vector embeddings using the relation `hasVector'.}
\label{fig:cyberexample}
\end{figure*}

By using the VKG structure, we link entity vector embeddings with their knowledge graph nodes. Domain specific knowledge graphs are built using a schema that is generally curated by domain experts. When we link the nodes and embeddings we can use the explicit information present in these ontologies to provide domain understanding to embeddings in vector space. Adding domain knowledge to vector embeddings can further improve various applications built upon the structure. The vector embeddings can be used to train machine learning models for various tasks. These machine learning models can use explicit assertions present in the knowledge graph part. This will also help in improving the quality of the results generated for various input queries discussed in Section \ref{query}.

%In another example from the medical domain (Figure \ref{fig:domain}) we use the VKG structure to state that the vector embedding of `skin' belongs to the class `Organ' which is a subclass of `Body Substance'. In a cybersecurity example (Figure \ref{fig:cyberexample}) from our Cyber-All-Intel system discussed in Section \ref{mysystem}, we explicitly state that `denial\_of\_service' is a vulnerability. Using the VKG structure we connect the fact that the vector embedding of `denial\_of\_service' is a vulnerability. 

In the example, (Figure \ref{fig:ex}) we can use the VKG structure to state that the vector embedding of `Milo' belongs to the class `Cat', which is a subclass of `Mammals' and so on. In a cybersecurity example (Figure \ref{fig:cyberexample}) from our Cyber-All-Intel system discussed in Section \ref{mysystem}, we explicitly state that `denial\_of\_service' is a vulnerability. Using the VKG structure we connect the fact that the vector embedding of `denial\_of\_service' is a vulnerability.

Knowledge graph nodes in the VKG structure can be used to add information from other sources like DBpedia, YAGO, and Freebase. This helps integrate information that is not present in the input corpus. For example, in Figure \ref{fig:cyberexample} we can link using the `$owl:sameAs$' property `Microsoft\_Internet\_Explorer' to its DBpedia equivalent `dbp:Internet\_Explorer'. Asserting this relation adds information like Internet Explorer is a product from Microsoft. This information may not have been present in the input cybersecurity corpus but is present in DBpedia.

In the rest of this section, we will discuss populating the VKG structure (Section \ref{pop}), query processing on the structure (Section \ref{query}).%, and various use-cases (Section \ref{apps}). 

\subsection{Populating the VKG Structure} \label{pop}

In order to create the VKG structure, the structure population system requires as input, a text corpus. The aim of the system is to create the VKG structure for the concepts and entities present in the input corpus which requires us to create the knowledge graph and the vector parts separately and then linking the two. 

Various steps and technologies required are enumerated below. The Cyber-All-Intel system that uses the VKG structure to represent a textual corpus is described in Section \ref{mysystem}.

\begin{enumerate}
\item \textit{Training vectors}: For the vector part of the VKG structure, we can generate entity embeddings using various vectorization algorithms that have been proposed. For text, many of these are based on the `Relational Concurrence Assumption' principle \cite{mikolov2013distributed,mikolov2013efficient,pennington2014glove,huang2012improving}.

\item \textit{Creating semantic triples}: In order to create triples for a textual corpus we need to extract entities using Named Entity Recognizers like, Stanford NER \cite{finkel2005incorporating}, extract relations between entities using relationship extractors like, Stanford openIE \cite{angeli2015leveraging}. After extracting entities and relationships these are asserted as RDF triples using domain specific ontologies.

\item \textit{Creating links between entity vectors and nodes}: After creating both the vector space model and the knowledge graph part, we link the node in the knowledge graph to the corresponding word in the vector space model's vocabulary using the $hasVector$ relationship shown in Figure \ref{fig:ex}. Keeping the vocabulary word in the knowledge graph allows us to update vector embeddings, if the underlining corpus gets updated or changed. This design decision was taken as the vector models go stale after some time due to the fact that the baseline input corpus on which these were trained change as new data is collected or retrieved in huge volume and with high velocity. The mapping between the vocabulary word and it's embedding is kept in the model output generated while training the vectors. This symbolic linking of the knowledge graph part and the vector part via the $hasVector$ relation is initiated after the RDF triples get generated. 
\end{enumerate}

The computational complexity to create the VKG structure can be derived form the steps mentioned above. These are detailed in extent literature \cite{mikolov2013efficient,finkel2005incorporating,angeli2015leveraging}. To create the vector space model the complexity is proportional to the word context window ($N$), vector dimensionality ($D$) and the hidden layer size ($H$)\footnote{Here we assume there is only 1 hidden layer}. To create the RDF triples, we use a named entity recognizer and relationship extractors where the complexity is proportional to the number of entity classes ($C$) and the set of relations ($R$). Computational complexity of the linking step depends on the vocabulary size of the corpus ($V$). So the computational complexity to create one structure: 

\[ (N \times D \times H)+(C+R)+V \]

While creating the VKG structures for the Cyber-All-Intel system on cybersecurity corpus of vocabulary size 246,321, the RDF triple generation part $(C + R)$ took most of the time. For an application the vector model may need to be retrained after some time to account for new incoming data, so the $(N \times D \times H)$ part runs periodically (in the Cyber-All-Intel system once every two weeks). Complexity of the linking process $(V)$ can be improved using hashing, it will also improve the performance of the query processing engine described in Section \ref{query}.

\begin{figure}[ht]
\centering
\includegraphics[scale=0.28]{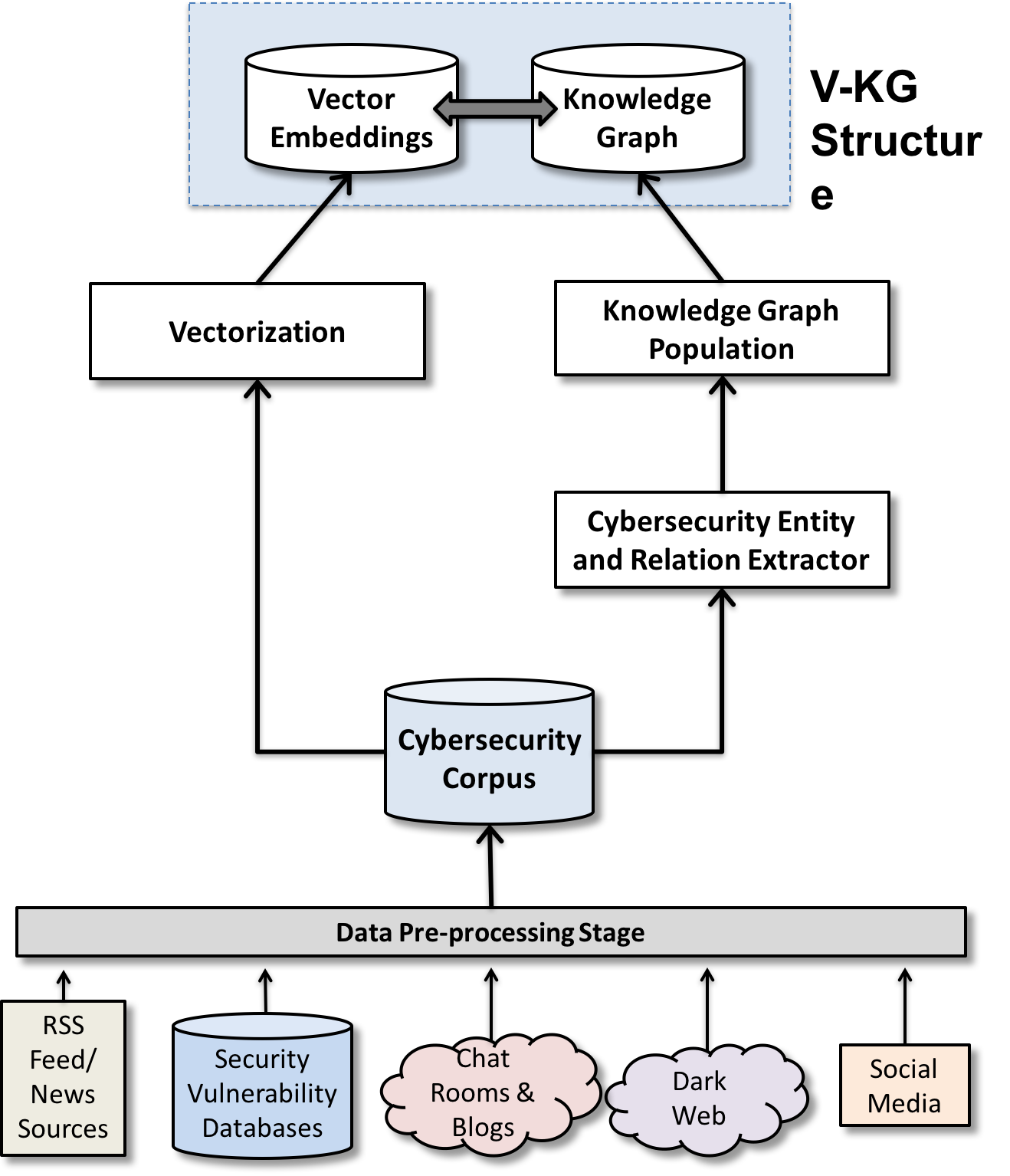}
\caption{C\MakeLowercase{yber}-A\MakeLowercase{ll}-I\MakeLowercase{ntel} System Architecture.}
\label{fig:arch}
\end{figure}

\subsubsection{C\MakeLowercase{yber}-A\MakeLowercase{ll}-I\MakeLowercase{ntel} System}\label{mysystem}

We created a multi-sourced cybersecurity threat intelligence system based on the VKG structure. A lot of data about potential attacks and various security vulnerabilities is available on-line in a multitude of data sources. Some of this data underlying this knowledge is in textual sources traditionally associated with Open Sources Intelligence (OSINT) \cite{osint}. The Cyber-All-Intel system (Figure \ref{fig:arch}) automatically accesses data from some of these sources like NIST's National Vulnerability Database, Twitter, Reddit\footnote{\url{https://www.reddit.com/}}, Security blogs, dark web markets \cite{dnmArchives}, etc. The system begins by collecting data in a modular fashion from these sources (see Figure \ref{fig:arch}). Followed by a data pre-processing stage where we remove stop words, perform stemming, noun chunking, etc. The data is then stored in a cybersecurity corpus. The next stages involve creating RDF triples and vector embeddings followed by linking.

The entities and relations for the knowledge graph part of the VKG structure are extracted by using a system similar to the one created by Joshi et al. \cite{Joshi-ICSC-2013}. The data is asserted in RDF using the Unified Cybersecurity Ontology (UCO) \cite{syed2015uco}. Ontologies like UCO, Intelligence \cite{mittal2016cybertwitter}, DBpedia \cite{auer2007dbpedia}, YAGO \cite{suchanek2007yago} have been used to provide cybersecurity domain knowledge. The vector part on the other hand was created using word2vec \cite{mikolov2013efficient}. An example is shown in Figure \ref{fig:cyberexample}, where we create the VKG structure for the textual input: 

\vspace{1mm}
\textit{Microsoft Internet Explorer allows remote attackers to execute arbitrary code or cause a denial of service (memory corruption) via a crafted web site, aka ``Internet Explorer Memory Corruption Vulnerability.''}
\vspace{1mm}

%\input{Sections/03_myrdf}

%Triples generated for the above mentioned input text are shown in the Figure \ref{fig:exampleRDF}. 
The knowledge part of the VKG structure is completed once the triples are added to the system. We used Apache Jena\footnote{\url{https://jena.apache.org/index.html}} to store our knowledge graph. For example, in Figure \ref{fig:cyberexample} once added the nodes are linked to the vector embeddings for `Microsoft\_Internet\_ Explorer', `remote\_attackers', `execute\_arbitrary \_code', `denial\_of\_ service', and `crafted\_web\_site'. For our system, we retrain the vector model every two weeks to incorporate the changes in the corpus. We give various details about system execution and evaluation in Section \ref{eval}.

\subsection{Query Processing}\label{query}

An application running on the VKG structure described in, Section \ref{mysystem} and populated via steps mentioned in Section \ref{pop}, can handle some specific type of queries. The application can ask a backend query processing engine to list declared
entities or relations, search for semantically similar concepts, and compute an output by reasoning over the stored data. This gives us three types of queries, $search$, $list$, and $infer$. These three are some of the basic tasks that an application running on the VKG structure will require, using which we derive our set of query commands ($C$):
\[ C = \{search,\, list,\, infer\} \]

A complex query posed by the application can be a union of some of these basic commands. An example query, to the Cyber-All-Intel security informatics system built on our VKG structure can be `list vulnerabilities in products similar to Google Chrome'; In this query we first have to $search$ for similar products to Google Chrome and then $list$ vulnerabilities found in these products. In a more general setting, an input query can be `Find similar sites to Taj Mahal, infer their distance from New York'; this includes a $search$ to find similar sites and to $infer$ their distance from New York. 

In the VKG structure, knowledge is represented in two parts, in a knowledge graph and in vector space. We argue that a query processing engine developed over the VKG structure  should combine the complimentary strengths of knowledge graphs and vector space models. Sub-queries involving similarity search-based tasks will give better and faster results, when carried out on the entire structure. Sub-queries that require declarative information retrieval, inference or reasoning should be carried out on the knowledge graph part. % (Table \ref{tab:op}). 

In both example queries mentioned above, $search$ queries, for the top-k nearest neighborhood search should be performed using the embeddings, and the $list$, $infer$ queries on the knowledge graph part. We evaluate the performance of these queries on different parts in Section \ref{eval}. 

%\begin{table}[h]
%\centering
%    \begin{tabular}{ccc}
%    \hline
%    \textbf{Operation}              & \textbf{Vectors}    & \textbf{ Graphs} \\ \hline
%    \vspace{1mm}
%    Similarity Search ($search$)     & \cmark & Aid vectors \\
%    \vspace{1mm}
%    Accessing declarations ($list$) & \xmark & \cmark \\
%    \vspace{1mm}
%    Reasoning ($infer$)             & \xmark & \cmark \\ \hline
%    \end{tabular}
%    \caption{Query Operations}
%\label{tab:op}
%\end{table}

Queries most suited for the vector part are those measuring semantic similarity of entities present in the corpus. One of the most simple similarity measures to compute entity similarity is by using the cosine of the angle between entity embeddings. Other query types include scoring, indexing, entity weighting, analogical mapping, nearest / neighborhood search, etc. Turney et al. provide a good survey of various applications and queries that can be built on vector space models \cite{turney2010frequency}. 

Some form of queries that can't be handled by the vector part are the ones that involve robust reasoning, these include any query that needs to infer logical consequences from a set of asserted triples using inference rules. Such queries need knowledge graph and rule engine support.  An argument can be made that in many applications it is enough to come up with a top-K plausible answer set using vector embeddings, however in expert systems like Cyber-All-Intel coming up with more concrete and well reasoned solutions is necessary. 

For the Cyber-All-Intel system described in Section \ref{mysystem}, some other example queries to the vector part can be, `Find products similar to Google Chrome.', `List vulnerabilities similar to buffer overflow', etc. 

Another advantage provided by integrating vector spaces and knowledge graphs is that we can use both of them to improve the results provided by either of the parts alone. For example (in Figure \ref{fig:cyberexample}), we can use the explicit information provided in the knowledge graph to aid the similarity search in vector space. If we are searching the vector space for entities similar to `denial\_of\_service', we can further improve our results by ensuring the entities returned belong to class `Vulnerability'. This information is available from the knowledge graph. This technique of knowledge graph aided vector space similarity search (\textit{VKG Search}) is used in our query engine. We execute similarity search on the embeddings and then filter out entities using the knowledge graph.

Type of queries that are well suited for knowledge graphs include querying the asserted facts for exact values of a triple's subject, predicate, or object. This information is not present in the vector part explicitly. Knowledge graphs also support an important class of queries that involve semantic reasoning or inference tasks based on rules that can be written in languages like SWRL. Domain experts can incorporate various reasoning and inference based techniques in the ontology for the knowledge graph part of the VKG structure. Mittal et al. in their system, CyberTwitter \cite{mittal2016cybertwitter} have showcased the use of an inference system to create threat alerts for cybersecurity using Twitter data. Such inference and reasoning tasks can be run on the knowledge graph part of the VKG structure. Knowledge graphs have been used to create various reasoning system where the reasoning logic is provided by the system creator. Hence, we created the $infer$ query which can be used to run application specific reasoning logic as in when required by the input query.

Knowledge graphs also provide the ability to integrate multiple sources of information. We can use the `$owl:sameAs$' relation to integrate other knowledge graphs. Once added these triples and reasoning techniques can be included in our $list$ and $infer$ queries. 

In the Cyber-All-Intel system the knowledge graph can handle queries like `What vulnerabilities are present in Internet Explorer?', `What products have the vulnerability buffer overflow?', etc. Quality of the output can be improved by running the similarity search on the vector spaces and using the knowledge graph to filter out entities not related to the input entity.

\textit{Adding to SPARQL:} Our query processing engine aims at extending SPARQL. In SPARQL, users are able to write `key-value' queries to a database that is a set of `subject-predicate-object' triples. Possible set of queries to SPARQL are, $Select$, $Construct$, $Ask$, $Describe$, and various forms of $Update$ queries. We create a layer above SPARQL to help integrate vector embeddings using our VKG structure. Our query processing engine sends $search$ queries to the vector part of the structure, the $list$ query to the SPARQL engine for the knowledge graph, and the $infer$ query to the Apache Jena inference engine. Next, we go into the details of our backend query processing system.

\subsubsection{Query processing system} Let a query proposed by an application to the backend system on the VKG structure be represented by $Q^{VKG}$. The task of the query processing engine is to run the input query, $Q^{VKG}$, as efficiently as possible. We evaluate this claim of efficiency in Section \ref{eval}. We do not discuss a query execution plan as multiple expert plans can be generated by domain experts depending on the needs of the application.

As per our need, in the backend system, a query that runs only on the knowledge graph part and only the vector part of the structure are represented as $Q^{kg}$ and $Q^{v}$ respectively. An input query $Q^{VKG}$ can be decomposed to multiple components that can run on different parts namely the knowledge graph and the vector part: 
\[ Q^{VKG} \rightarrow Q^{kg} \cup Q^{v} \]

An input application query $Q^{VKG}$ can have multiple components that can run on the same part, for example, an input query can have three components, two of which run on the knowledge graph part and the remaining one runs on the vector part. Such a query can be represented as: 
\[ Q^{VKG} \rightarrow Q^{v} \cup Q_{1}^{kg} \cup Q_{2}^{kg} \tag{1}\label{q1}  \]

Where, $Q_{1}^{kg}$ and $Q_{2}^{kg}$ are the two components that run on the knowledge graph part and $Q^{v}$ component which runs on the vector part. 

It is the responsibility of the query processing system to execute these subqueries on different parts and combine their output to compute the answer to the original input query $Q^{VKG}$. We describe the query execution process using an example. 

\subsubsection{Example query} For the Cyber-All-Intel system an example query issued by the application: `Infer an alert if, a vulnerability similar to denial of service is listed in MySQL', can be considered as three sub-queries which need to be executed on different parts of the VKG structure.

The input query can be considered to be of the type \eqref{q1}, Where the subqueries are:
\begin{enumerate}
\item Finding similar vulnerabilities (set - $V$) to denial of service that will run on the vector embeddings ($Q^{v}$).
\item Listing known existing vulnerabilities (set - $K$) in MySQL ($Q_{1}^{kg}$).
\item Inferring if an alert should be raised if a vulnerability (from set $V$) is found in the product MySQL. This sub-query will run on the knowledge graph part ($Q_{2}^{kg}$).
\end{enumerate}
The query can be represented as: 
\begin{multline*}
Q^{VKG} = \{\{search, \, `denial\_of\_service', \, V\} \,\,\cup \\
\{list, vulnerability, `MySQL', \, K\} \,\,\cup \\
\{infer, V, K, `MySQL', alert\}\} \tag{Query 1}\label{q2} 
\end{multline*}

The query execution plan for \eqref{q2} is to first run $Q^{v}$ and $Q_{1}^{kg}$ simultaneously and compute the sets $V$ and $K$. After computing the sets the engine is supposed to run $Q_{2}^{kg}$.

The first part of the input query \eqref{q2}, is of the form $Q^{v}$ and will run on the vector part of the VKG structure. Its representation is: 
 \[ Q^{v} = \{search, \, `denial\_of\_service', \, V\} \]

\begin{figure}[ht]
\centering
\includegraphics[scale=0.2]{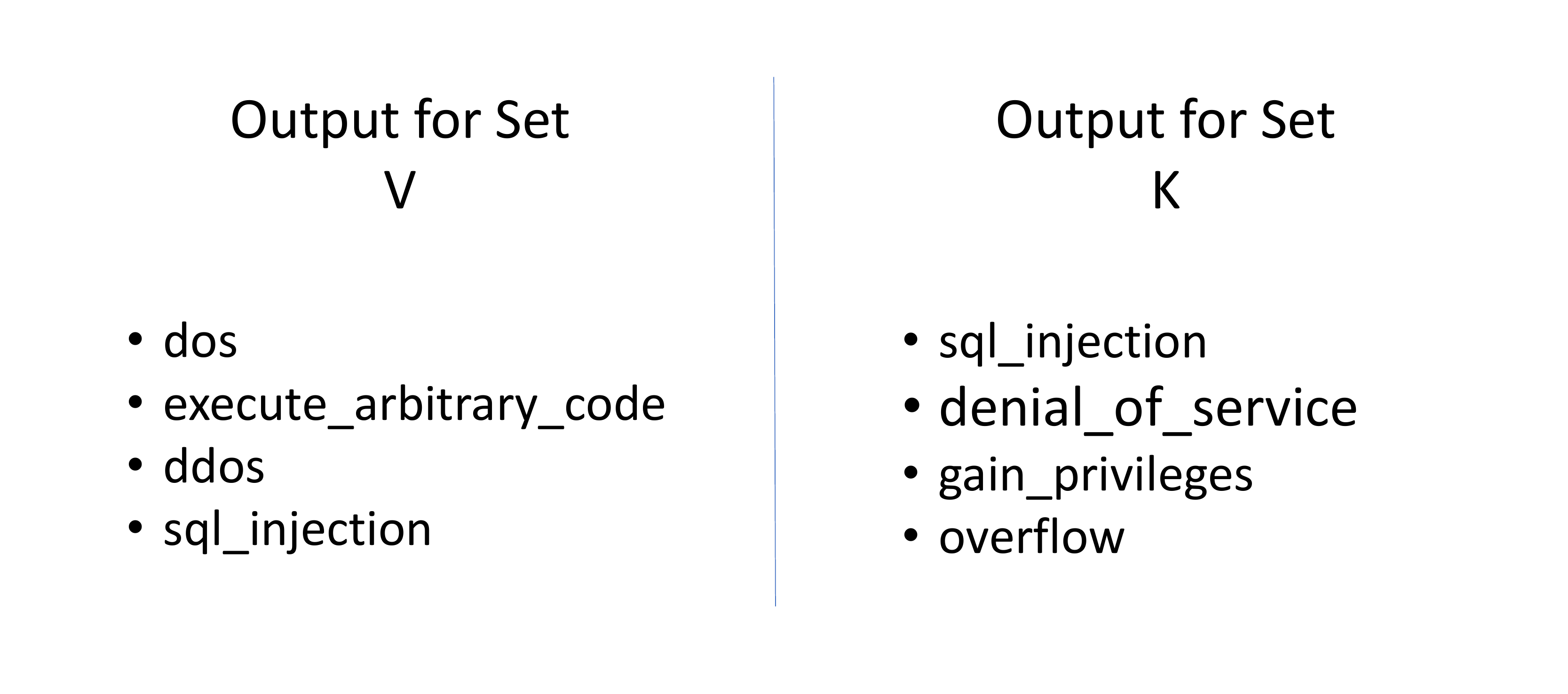}
\caption{The output of the sub-queries $Q^{v}$ and $Q_{1}^{kg}$ when run on the Cyber-All-Intel System. As there is some overlap between the sets $V$ and $K$ the output for the subquery $Q_{2}^{kg}$ will be `Alert = Yes'}
\label{fig:qo}
\end{figure}

The output generated is a set $V$ (Figure \ref{fig:qo}) and contains vulnerabilities similar to `denial\_of\_service'. We used the VKG search to compute this set and filter out all non vulnerabilities. The set $V$ will be utilized by other subqueries ($Q_{2}^{kg}$) to generate it's output. 

The second part of the input query \eqref{q2} is the first sub-query to run on the knowledge graph part of the VKG structure. 
\[ Q_{1}^{kg} = \{list, vulnerability, `MySQL', \, K\}\]

The goal of this query is to list all vulnerabilities (Figure \ref{fig:qo}) present in `MySQL' that are explicitly mentioned in the knowledge graph (set - $K$).

The third part of the input query \eqref{q2} is the second subquery to run on the knowledge graph part of the VKG structure. 
\[ Q_{2}^{kg} = \{infer, V, K, `MySQL', alert\} \]

Here, the output is to reason whether to raise an `alert' if some overlap is found between the sets $V$ \& $K$. Query $Q_{2}^{kg}$ requires an inference engine to output an alert based on some logic provided by domain experts or system security administrators. In Figure \ref{fig:qo} as there is overlap between the sets $V$ and $K$ an alert will be raised. 

\section{Experimental Results}\label{eval}

We describe the experimental setup, evaluation and results obtained by running the Cyber-All-Intel system. We first go through the data description and then evaluate the VKG structure and the query processing engine using various evaluation measures and runtime results.

\subsection{Dataset and Experimental Setup}

For Cyber-All-Intel system, we created a Cybersecurity corpus as discussed in Section \ref{mysystem} and shown in Figure \ref{fig:arch}. Data for the corpus is collected from many sources, including chat rooms, dark web, blogs, RSS feeds, social media, and vulnerability databases. The current corpus has 85,190 common vulnerabilities and exposures from the NVD dataset maintained by the MITRE corporation, 351,004 cleaned Tweets collected through the Twitter API, 25,146 Reddit and blog posts from sub-reddits like, r/cybersecurity/, r/netsec/, etc. and a few dark web posts \cite{dnmArchives}. 

For the vector space models, we created word2vec embeddings by setting vector dimensions as: 500, 1000, 1500, 1800, 2500 and term frequency as: 1, 2, 5, 8, 10 for each of the dimensions. The context window was set at 7. The knowledge graph part was created using the the steps mentioned in Section \ref{mysystem} and the VKG structure was generated by linking the knowledge graph nodes with their equivalents in the vector model vocabulary (see Section \ref{pop}).

In order to conduct various evaluations, we first created an annotated test set. We selected some data from the cybersecurity corpus and had it annotated by a group of five graduate students familiar with cybersecurity concepts. The annotators were asked to go through the corpus and mark the following entity classes: \textit{Address, Attack / Incident, Attacker, Campaign, Attacker, CVE, Exploit, ExploitTarget, File, Hardware, Malware, Means, Consequence, NetworkState, Observable, Process, Product, Software, Source, System, Vulnerability, Weakness}, and \textit{VersionNumber}. They were also asked to annotate various relations including \textit{hasAffectedSoftware, hasAttacker, hasMeans, hasWeakness, isUnderAttack, hasSoftware, has CVE\_ID}, and \textit{hasVulnerability}. These classes and relations correspond to various classes and properties listed in the Unified Cybersecurity Ontology and the Intelligence Ontology \cite{mittal2016cybertwitter}. For the annotation experiment, we computed the inter-annotator agreement score using the Cohen's Kappa \cite{10.2307/2531300}. Only the annotations above the agreement score of 0.7 were kept.

The annotators were also tasked to create sets of similar products and vulnerabilities so as to test various aspects of the Cyber-All-Intel system. The most difficult task while designing various experiments and annotation tasks was to define the meaning of the word `similar'. Should similar products have the same vulnerabilities, or same use? In case of our cybersecurity corpus we found that the two sets, same vulnerabilities and same use were co-related. For example, if two products have SQL injection vulnerability we can say with certain confidence that they use some form of a database technology and may have similar features and use. If they have Cross-Site Request Forgery (CSRF) vulnerability they may generally belong to the product class of browsers.

Annotators manually created certain groups of products like, operating systems, browsers, databases, etc. OWASP\footnote{\url{https://www.owasp.org/index.php/Main_Page}} maintains groups of similar vulnerabilities\footnote{\url{https://www.owasp.org/index.php/Category:Vulnerability}} and attacks\footnote{\url{https://www.owasp.org/index.php/Category:Attack}}. We created 14 groups of similar vulnerabilities, 11 groups of similar attacks, 31 groups of similar products. A point to note here is that, in many cases certain entities are sometimes popularly referred by their abbreviations, we manually included abbreviations in these 56 groups. For example, we included DOS and CSRF which are popular abbreviations for Denial Of Service and Cross-Site Request Forgery respectively in various groups.

\subsection{Evaluations}

Using the above mentioned data and annotation test sets we evaluate the different parts of our VKG structure and the query processing engine. In Section \ref{query}, we describe our query processing engine and its three query commands: $search$, $list$, and $infer$. Here we evaluate $search$ and $list$ query but not the $infer$ queries as they depend on the reasoning logic provided in the ontology and can vary with application.

\textit{Evaluation tasks:} We evaluate $search$ by comparing the performance of three models: vector embeddings, knowledge graph, and VKG search. Queries of type $list$ can only be run on the knowledge graph part of the VKG structure as there is no declarative information present in the embeddings. To evaluate the $list$ query we evaluate the quality of triples generated in the knowledge graph. We also evaluate the vector space and knowledge graph linking, as good quality linking is necessary for various operations on the VKG structure.

\subsubsection{Evaluating the $search$ query}

An input query to the VKG structure to find similar concepts can either run on the vector space using various neighborhood search algorithms \cite{gionis1999similarity,Kuzi:2016:QEU:2983323.2983876} or the knowledge graph using ontology matching, ontology alignment, schema matching, instance matching, similarity search, etc. \cite{shvaiko2013ontology,DeVirgilio:2013:SMA:2457317.2457352,zheng2016semantic}. We evaluate claims in Section \ref{query} that we get better and faster results using the vector space instead of the knowledge graph for similarity computations. We also show that knowledge graph aided vector spaces (VKG search) provide the best results when compared to vector spaces and knowledge graphs alone.  

%In order to evaluate the $search$ query we evaluate the vector space part and compare searching on the vector space to instance matching on the knowledge graph. 

\begin{figure}[h]
\centering
\includegraphics[scale=0.25]{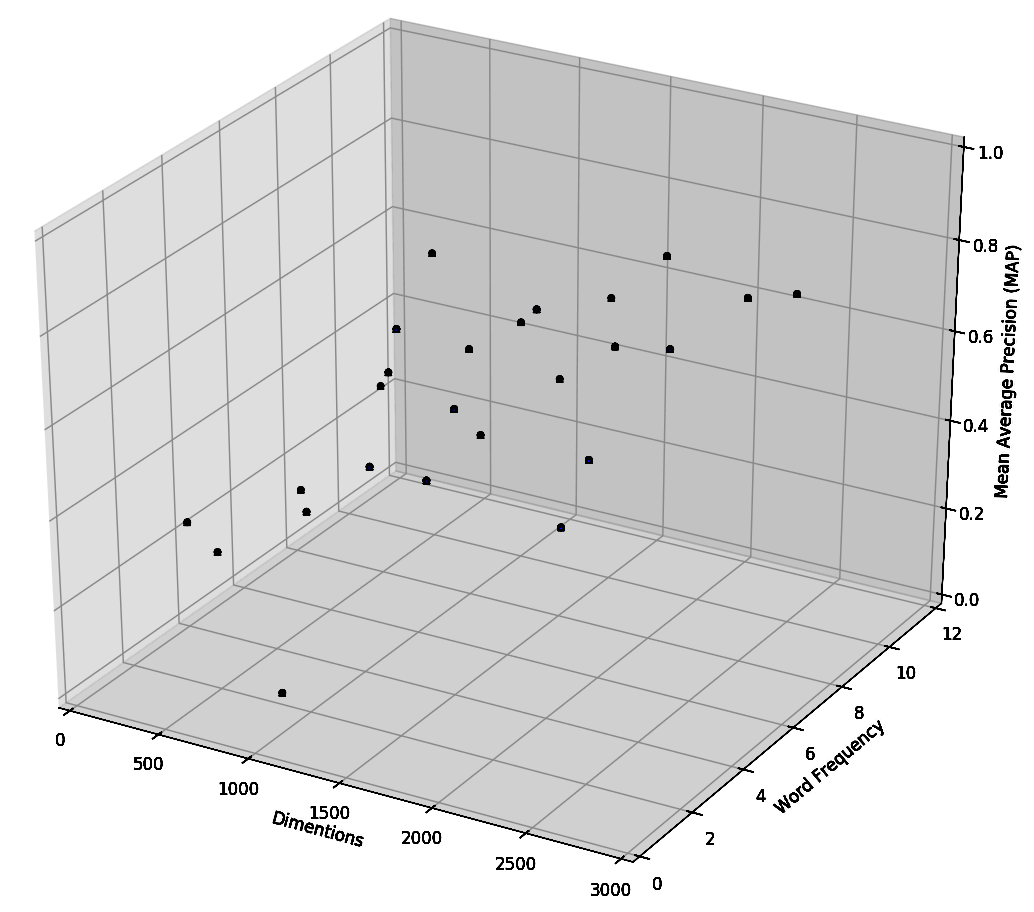}
\caption{Mean Average Precision for different dimensions and word frequency. Models with higher dimensions and word frequency performed better.}
\label{fig:map}
\end{figure}

\textit{Evaluating the Vector Space Models:} To evaluate the vector embeddings part of the VKG structure we used the `similar' sets created by the annotators. We trained various vector space models with vector dimension, 500, 1000, 1500, 1800, 2500 and term frequency, 1, 2, 5, 8, 10. Increasing the value of dimensionality and decreasing the term frequency almost exponentially increases the time to generate the vector space models. We first find the combination of parameters for which the Mean Average Precision (MAP) is highest, so as to use it in comparing the performance of vector space models with knowledge graphs and graph aided vectors in the VKG structure, in finding similar vulnerabilities, attacks, and products. For the 56 similar groups the vector model with dimensionality of 1500 and term frequency 2, had the highest MAP of 0.69 (Figure \ref{fig:map}). Models with higher dimensions and word frequency performed better.

\begin{figure}[h]
\centering
\includegraphics[scale=0.17]{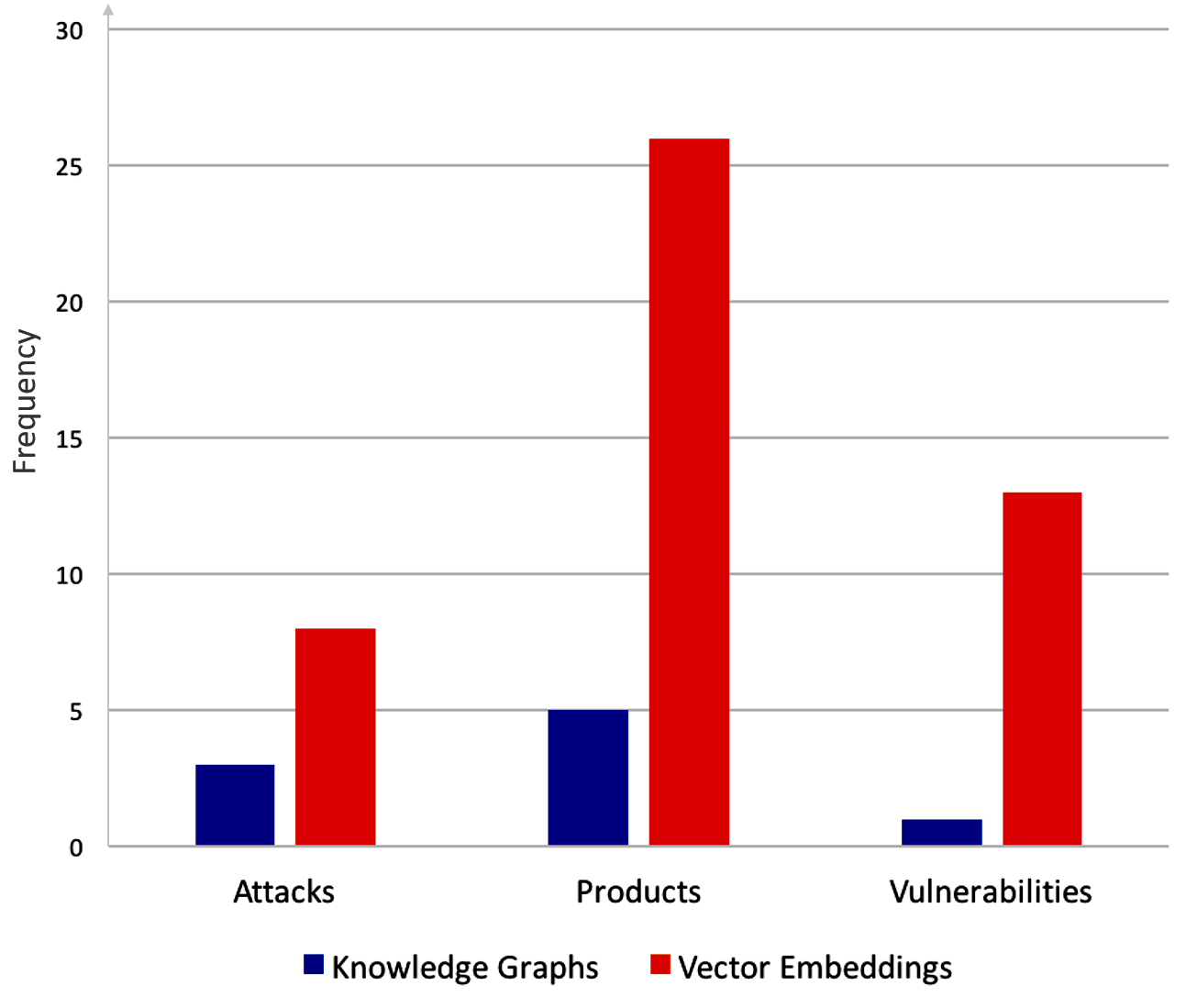}
\caption{The number of times the MAP score was higher for the two knowledge representation techniques for the 56 similar groups. Vector embeddings performed better than knowledge graphs. Embeddings performed better in 8 attacks, 26 products, and 13 vulnerabilities.}
\label{fig:map2}
\end{figure}

\textit{Comparing Vector Space Models and Knowledge Graphs:}  To compare the performance of the $search$ query over vector space and its counterpart from the knowledge graph side we used the vector embedding model with dimensionality of 1500 and term frequency 2. To compute instance matching on knowledge graphs, we used an implementation of ASMOV (Automated Semantic Matching of Ontologies with Verification) \cite{JeanMary2009OntologyMW}. On computing the MAP for both vector embeddings and the knowledge graphs we found that embeddings constantly outperformed the knowledge graphs. Figure \ref{fig:map2}, shows that the MAP value for vector embeddings was higher 47 times out of 56 similarity groups considered. The knowledge graph performed significantly bad for vulnerabilities and attacks as the structural schema for both attacks and vulnerabilities was quite dense with high number of edges to different entities. This significantly affected the performance of schema matching. We also found that computing $search$ queries was faster on vector embeddings as compared to knowledge graphs. Neighborhood search on vector spaces was empirically 11 times faster in our experiments compared to knowledge graphs.

\textit{Comparing Vector Space Models and VKG Search:} To test the advantages of our VKG structure we evaluate the VKG search (see Section \ref{query}) against the vector space model. The VKG search on vector space achieved a MAP of 0.8, which was significantly better than the MAP score (0.69) achieved by using just the vector model. The reason for higher quality results obtained by using the VKG search is due to the fact that we can filter out entities by using class type declarations present in the knowledge graph.

\begin{table}[h]
\centering
    \begin{tabular}{|c|c|c|c|}
    \hline
    \textbf{Model} & Graphs & Vectors & VKG Search \\ \hline
    \textbf{MAP}   & 0.43   & 0.69    & 0.80         \\ \hline
    \end{tabular}
    \caption{Best Mean Average Precision for knowledge graphs, vector space models, and VKG structure.}
\end{table}

\subsubsection{Evaluating the $list$ query}

Since the declarative assertions are made in the VKG's knowledge graph, to evaluate the $list$ query we evaluate the quality of the knowledge graph part. The $list$ query can not be executed on the vector space as there is no declarative information in embeddings. 

To check the quality of the knowledge graph triples generated from the raw text we asked the same set of annotators to manually evaluate the triples created and compare them with the original text. The annotators were given three options, correct, partially correct, and wrong. From 250 randomly selected text samples from the cybersecurity data, the annotators agreed that 83\% were marked correct, 9\% were partially correct, and 8\% were marked wrong.

\subsubsection{Evaluating Vector and Graph linking} To check the quality of the linkages created in the VKG structure, we again asked the annotators to review various nodes in the triples and check if they were linked to the right vocabulary word from the vector models generated. On evaluating 250 randomly selected triples the annotators found 97\% of them were linked correctly. This high accuracy can be attributed to the fact that both knowledge graphs and the vector space models had the same underlying corpus vocabulary.

\section{Conclusion \& Future Work}
\label{conc}

Vector space models and knowledge graph both have some form of information loss, vector embeddings tend to lose the declarative nature of information, but are ideal for similarity computation, indexing, scoring, neighborhood search, etc. Knowledge graphs, on the other hand, are good for asserting declarative information and have well developed reasoning techniques, perform slower. Together, they can both add value to each other. Using a hybrid of the two we can use the knowledge graph's declarative powers to explicitly state relations between various embeddings in a vector space. For example, we can assert in a knowledge graph, the information that `an adult mouse has 16 teeth' which is tough to do in a vector space. On the other hand we can use the embeddings to figure out entities similar to `Arctic Tern' and use the knowledge graph to improve the quality of our results, by understanding that the query pertains to various bird species. 

In this paper, we described the design and implementation of a VKG structure which is part vector embeddings and part knowledge graph. We discussed a method to create the VKG structure and built a query processing engine that decomposes an input query to sub-queries with commands $search$, $list$, and $infer$. We showed that queries of the type $search$ should run on the vector part of the VKG structure and the results should be improved with the aid of knowledge graph assertions. We call this technique VKG search. $list$ and $infer$ run on the knowledge graph part to take advantage of the robust reasoning capabilities. Using the VKG structure, we can efficiently answer those queries which are not handled by vector embeddings or knowledge graphs alone.

We also demonstrated the VKG structure and the query processing engine by developing a system called \textit{Cyber-All-Intel} for knowledge extraction, representation and querying in an end-to-end pipeline grounded in the cybersecurity informatics domain. The system creates a cybersecurity corpus by collecting threat and vulnerability intelligence from various textual sources like, national vulnerability databases, dark web vulnerability markets, social networks, blogs, etc. which are represented as instances of our VKG structure. We use the system to answer complex cybersecurity informatics queries. 

In our future work, we plan to add more types of commands and queries that can be processed on our VKG structure. We would also like to build more use-cases and applications on the VKG structure.%, some of which were mentioned in Section \ref{apps}. 

%input as a textual command 
%textit{Creating vectors for derived concepts:} In the VKG structure the knowledge graph can have nodes that are generated for schema requirements like the `Int3482758232' node created in Figure \ref{fig:exampleRDF} and Figure \ref{fig:cyberexample}. The `Int3482758232' represents the intelligence node in the structure created to represent a particular input intelligence. The vector representation for the `Int3482758232' node can be functionally derived using the vectors of nodes execute\_arbitrary\_code, denial\_of\_service, Microsoft\_Internet\_Explorer, crafted\_web\_site, remote\_ attackers, etc.

\bibliographystyle{IEEEtran}
\bibliography{phd}

\begin{thebibliography}{10}
\providecommand{\url}[1]{#1}
\csname url@samestyle\endcsname
\providecommand{\newblock}{\relax}
\providecommand{\bibinfo}[2]{#2}
\providecommand{\BIBentrySTDinterwordspacing}{\spaceskip=0pt\relax}
\providecommand{\BIBentryALTinterwordstretchfactor}{4}
\providecommand{\BIBentryALTinterwordspacing}{\spaceskip=\fontdimen2\font plus
\BIBentryALTinterwordstretchfactor\fontdimen3\font minus
  \fontdimen4\font\relax}
\providecommand{\BIBforeignlanguage}[2]{{%
\expandafter\ifx\csname l@#1\endcsname\relax
\typeout{** WARNING: IEEEtran.bst: No hyphenation pattern has been}%
\typeout{** loaded for the language `#1'. Using the pattern for}%
\typeout{** the default language instead.}%
\else
\language=\csname l@#1\endcsname
\fi
#2}}
\providecommand{\BIBdecl}{\relax}
\BIBdecl

\bibitem{davis1993knowledge}
R.~Davis, H.~Shrobe, and P.~Szolovits, ``What is a knowledge representation?''
  \emph{AI magazine}, vol.~14, no.~1, p.~17, 1993.

\bibitem{mikolov2013efficient}
T.~Mikolov, K.~Chen, G.~Corrado, and J.~Dean, ``Efficient estimation of word
  representations in vector space,'' \emph{preprint arXiv:1301.3781}, 2013.

\bibitem{pennington2014glove}
J.~Pennington, R.~Socher, and C.~D. Manning, ``Glove: Global vectors for word
  representation.'' in \emph{EMNLP}, vol.~14, 2014, pp. 1532--43.

\bibitem{kahneman2011}
D.~Kahneman, \emph{Thinking, fast and slow}.\hskip 1em plus 0.5em minus
  0.4em\relax Macmillan, 2011.

\bibitem{mikolov2013distributed}
T.~Mikolov and J.~Dean, ``Distributed representations of words and phrases and
  their compositionality,'' \emph{Advances in neural information processing
  systems}, 2013.

\bibitem{sutskever2014sequence}
I.~Sutskever, O.~Vinyals, and Q.~V. Le, ``Sequence to sequence learning with
  neural networks,'' in \emph{Advances in neural information processing
  systems}, 2014, pp. 3104--3112.

\bibitem{huang2012improving}
E.~H. Huang, R.~Socher, C.~D. Manning, and A.~Y. Ng, ``Improving word
  representations via global context and multiple word prototypes,'' in
  \emph{50th ACL}.\hskip 1em plus 0.5em minus 0.4em\relax ACL, 2012, pp.
  873--882.

\bibitem{auer2007dbpedia}
S.~Auer, C.~Bizer, G.~Kobilarov, J.~Lehmann, R.~Cyganiak, and Z.~Ives,
  \emph{DBpedia: A Nucleus for a Web of Open Data}.\hskip 1em plus 0.5em minus
  0.4em\relax Springer, 2007.

\bibitem{suchanek2007yago}
F.~M. Suchanek, G.~Kasneci, and G.~Weikum, ``Yago: a core of semantic
  knowledge,'' in \emph{Proceedings of the 16th international conference on
  World Wide Web}.\hskip 1em plus 0.5em minus 0.4em\relax ACM, 2007, pp.
  697--706.

\bibitem{hoffart2013yago2}
J.~Hoffart, F.~M. Suchanek, K.~Berberich, and G.~Weikum, ``Yago2: A spatially
  and temporally enhanced knowledge base from wikipedia,'' \emph{Artificial
  Intelligence}, vol. 194, pp. 28--61, 2013.

\bibitem{officialgoogle2012}
\BIBentryALTinterwordspacing
A.~Singhal, ``Introducing the knowledge graph: things, not strings,'' May 2012.
  [Online]. Available:
  \url{https://googleblog.blogspot.com/2012/05/introducing-knowledge-graph-things-not.html}
\BIBentrySTDinterwordspacing

\bibitem{bollacker2008freebase}
K.~Bollacker, C.~Evans, P.~Paritosh, T.~Sturge, and J.~Taylor, ``Freebase: a
  collaboratively created graph database for structuring human knowledge,'' in
  \emph{SIGMOD Int. Conf on Management of Data}.\hskip 1em plus 0.5em minus
  0.4em\relax ACM, 2008, pp. 1247--1250.

\bibitem{gionis1999similarity}
A.~Gionis, P.~Indyk, and R.~Motwani, ``Similarity search in high dimensions via
  hashing,'' in \emph{VLDB}, vol.~99, no.~6, 1999, pp. 518--529.

\bibitem{Kuzi:2016:QEU:2983323.2983876}
S.~Kuzi, A.~Shtok, and O.~Kurland, ``Query expansion using word embeddings,''
  in \emph{25th Int. Conf. on Information and Knowledge Management}.\hskip 1em
  plus 0.5em minus 0.4em\relax ACM, 2016, pp. 1929--1932.

\bibitem{shvaiko2013ontology}
P.~Shvaiko and J.~Euzenat, ``Ontology matching: state of the art and future
  challenges,'' \emph{IEEE Transactions on knowledge and data engineering},
  vol.~25, no.~1, pp. 158--176, 2013.

\bibitem{DeVirgilio:2013:SMA:2457317.2457352}
R.~De~Virgilio, A.~Maccioni, and R.~Torlone, ``A similarity measure for
  approximate querying over {RDF} data,'' in \emph{Proc. Joint EDBT/ICDT 2013
  Workshops}.\hskip 1em plus 0.5em minus 0.4em\relax ACM, 2013, pp. 205--213.

\bibitem{zheng2016semantic}
W.~Zheng, L.~Zou, W.~Peng, X.~Yan, S.~Song, and D.~Zhao, ``Semantic sparql
  similarity search over {RDF} knowledge graphs,'' \emph{Proceedings of the
  VLDB Endowment}, vol.~9, no.~11, pp. 840--851, 2016.

\bibitem{sleeman2015topic}
J.~Sleeman, T.~Finin, and A.~Joshi, ``Topic modeling for rdf graphs,'' in
  \emph{3rd Int. Workshop on Linked Data for Information Extraction}, Oct.
  2015, pp. 48--62.

\bibitem{yang2016fast}
S.~Yang, F.~Han, Y.~Wu, and X.~Yan, ``Fast top-k search in knowledge graphs,''
  in \emph{Data Engineering (ICDE), 2016 IEEE 32nd International Conference
  on}.\hskip 1em plus 0.5em minus 0.4em\relax IEEE, 2016, pp. 990--1001.

\bibitem{damljanovic2011random}
D.~Damljanovic, J.~Petrak, M.~Lupu, H.~Cunningham, M.~Carlsson, G.~Engstrom,
  and B.~Andersson, ``Random indexing for finding similar nodes within large
  {RDF} graphs,'' in \emph{Extended Semantic Web Conference}.\hskip 1em plus
  0.5em minus 0.4em\relax Springer, 2011, pp. 156--171.

\bibitem{nickel2016holographic}
M.~Nickel, L.~Rosasco, and T.~Poggio, ``Holographic embeddings of knowledge
  graphs.'' in \emph{AAAI}, 2016, pp. 1955--1961.

\bibitem{wang2014knowledge}
Z.~Wang, J.~Zhang, J.~Feng, and Z.~Chen, ``Knowledge graph embedding by
  translating on hyperplanes.'' in \emph{AAAI}.\hskip 1em plus 0.5em minus
  0.4em\relax Citeseer, 2014, pp. 1112--1119.

\bibitem{lin2015learning}
Y.~Lin, Z.~Liu, M.~Sun, Y.~Liu, and X.~Zhu, ``Learning entity and relation
  embeddings for knowledge graph completion.'' in \emph{AAAI}, 2015, pp.
  2181--2187.

\bibitem{neelakantan2015compositional}
A.~Neelakantan, B.~Roth, and A.~McCallum, ``Compositional vector space models
  for knowledge base completion,'' \emph{arXiv preprint arXiv:1504.06662},
  2015.

\bibitem{socher2013reasoning}
R.~Socher, D.~Chen, C.~D. Manning, and A.~Ng, ``Reasoning with neural tensor
  networks for knowledge base completion,'' in \emph{Advances in Neural
  Information Processing Systems}, 2013, pp. 926--934.

\bibitem{guu2015traversing}
K.~Guu, J.~Miller, and P.~Liang, ``Traversing knowledge graphs in vector
  space,'' \emph{arXiv preprint arXiv:1506.01094}, 2015.

\bibitem{Undercoffer2003b}
J.~Undercofer, A.~Joshi, and J.~Pinkston, ``{Modeling Computer Attacks: An
  Ontology for Intrusion Detection},'' in \emph{Proc. 6th Int. Symposium on
  Recent Advances in Intrusion Detection}.\hskip 1em plus 0.5em minus
  0.4em\relax Springer, September 2003.

\bibitem{kandefer2007symbolic}
M.~Kandefer, S.~Shapiro, A.~Stotz, and M.~Sudit, ``Symbolic reasoning in the
  cyber security domain,'' 2007.

\bibitem{takahashi2010ontological}
T.~Takahashi, Y.~Kadobayashi, and H.~Fujiwara, ``Ontological approach toward
  cybersecurity in cloud computing,'' in \emph{3rd Int. Conf. on Security of
  information and networks}.\hskip 1em plus 0.5em minus 0.4em\relax ACM, 2010,
  pp. 100--109.

\bibitem{takahashi2010building}
T.~Takahashi, H.~Fujiwara, and Y.~Kadobayashi, ``Building ontology of
  cybersecurity operational information,'' in \emph{6th Workshop on Cyber
  Security and Information Intelligence Research}.\hskip 1em plus 0.5em minus
  0.4em\relax ACM, 2010, p.~79.

\bibitem{rutkowski2010cybex}
A.~Rutkowski, Y.~Kadobayashi, I.~Furey, D.~Rajnovic, R.~Martin, T.~Takahashi,
  C.~Schultz, G.~Reid, G.~Schudel, M.~Hird \emph{et~al.}, ``Cybex: the
  cybersecurity information exchange framework (x.1500),'' \emph{SIGCOMM
  Computer Communication Review}, vol.~40, no.~5, pp. 59--64, 2010.

\bibitem{syed2015uco}
Z.~Syed, A.~Padia, M.~L. Mathews, T.~Finin, and A.~Joshi, ``{UCO}: A unified
  cybersecurity ontology,'' in \emph{AAAI Workshop on Artificial Intelligence
  for Cyber Security}.\hskip 1em plus 0.5em minus 0.4em\relax AAAI Press, 2015,
  pp. 14--21.

\bibitem{Joshi-ICSC-2013}
A.~Joshi, R.~Lal, T.~Finin, and A.~Joshi, ``{Extracting cybersecurity related
  linked data from text},'' in \emph{Proceedings of the 7th IEEE International
  Conference on Semantic Computing}.\hskip 1em plus 0.5em minus 0.4em\relax
  IEEE Computer Society Press, September 2013.

\bibitem{mittal2016cybertwitter}
S.~Mittal, P.~K. Das, V.~Mulwad, A.~Joshi, and T.~Finin, ``Cybertwitter: Using
  twitter to generate alerts for cybersecurity threats and vulnerabilities,''
  in \emph{IEEE/ACM Int. Conf. on Advances in Social Networks Analysis and
  Mining}.\hskip 1em plus 0.5em minus 0.4em\relax IEEE, 2016, pp. 860--867.

\bibitem{mendes2011dbpedia}
P.~N. Mendes, M.~Jakob, A.~Garc\'{\i}a-Silva, and C.~Bizer, ``{DBpedia
  spotlight: shedding light on the web of documents},'' in \emph{7th Int. Conf.
  on Semantic Systems}.\hskip 1em plus 0.5em minus 0.4em\relax ACM, 2011, pp.
  1--8.

\bibitem{collobert2011natural}
R.~Collobert, J.~Weston, L.~Bottou, M.~Karlen, K.~Kavukcuoglu, and P.~Kuksa,
  ``Natural language processing (almost) from scratch,'' \emph{Journal of
  Machine Learning Research}, vol.~12, no. Aug, pp. 2493--2537, 2011.

\bibitem{berners2001semantic}
T.~Berners-Lee, J.~Hendler, and O.~Lassila, ``The semantic web,'' \emph{The
  Scientific American}, 2001.

\bibitem{JeanMary2009OntologyMW}
Y.~R. Jean-Mary, E.~P. Shironoshita, and M.~R. Kabuka, ``Ontology matching with
  semantic verification,'' \emph{Jour. Web Semantics}, vol. 7 3, pp. 235--251,
  2009.

\bibitem{finkel2005incorporating}
J.~R. Finkel, T.~Grenager, and C.~Manning, ``Incorporating non-local
  information into information extraction systems by gibbs sampling,'' in
  \emph{43rd annual meeting on Association For Computational
  Linguistics}.\hskip 1em plus 0.5em minus 0.4em\relax ACL, 2005, pp. 363--370.

\bibitem{angeli2015leveraging}
G.~Angeli, M.~J. Premkumar, and C.~D. Manning, ``Leveraging linguistic
  structure for open domain information extraction,'' in \emph{53rd Annual
  Meeting of the Association for Computational Linguistics}, 2015.

\bibitem{osint}
R.~D. Steele, ``Open source intelligence: What is it? why is it important to
  the military,'' \emph{American Intelligence Journal}, vol.~17, no.~1, pp.
  35--41, 1996.

\bibitem{dnmArchives}
G.~Branwen, N.~Christin, D.~Décary-Hétu, R.~M. Andersen, StExo,
  E.~Presidente, Anonymous, D.~Lau, Sohhlz, D.~Kratunov, V.~Cakic, V.~Buskirk,
  and Whom, ``Dark net market archives, 2011-2015,''
  \url{https://www.gwern.net/DNM%20archives}, July 2015.

\bibitem{turney2010frequency}
P.~D. Turney and P.~Pantel, ``From frequency to meaning: Vector space models of
  semantics,'' \emph{Jour. Artificial Intelligence Research}, vol.~37, pp.
  141--188, 2010.

\bibitem{10.2307/2531300}
N.~C. Smeeton, ``Early history of the kappa statistic,'' \emph{Biometrics},
  vol.~41, no.~3, pp. 795--795, 1985.

\end{thebibliography}

\end{document}